\documentclass[journal]{IEEEtran}

\usepackage{url}
\usepackage{color}
\usepackage{amsmath}
\usepackage{array,graphicx,subfigure}
\usepackage{multirow}
\usepackage{amssymb}
\usepackage{bbding}

\usepackage{colortbl}

\usepackage{bbm}

\usepackage{amsfonts}
\usepackage{hyperref}
\hypersetup{
    colorlinks=true,
    linkcolor=black,
    filecolor=black,      
    urlcolor=black,
    citecolor=black,
}

\usepackage{algorithm}
\usepackage{algpseudocode}

\newcommand{\tabincell}[2]{\begin{tabular}{@{}#1@{}}#2\end{tabular}}

\newcommand{\cut}[1]{}

\newcommand{\todo}{\textcolor{red}{TODO}}

\newcommand{\ie}{\textit{i}.\textit{e}.}
\newcommand{\eg}{\textit{e}.\textit{g}.}
\newcommand{\etc}{\textit{etc}}

\ifCLASSINFOpdf
  
\else
 
\fi

\begin{document}

\title{On Learning Semantic Representations for Million-Scale Free-Hand Sketches}

\author{Peng~Xu$^{*}$,~Yongye Huang,~Tongtong Yuan,~Tao Xiang,\\
~Timothy M. Hospedales,~\IEEEmembership{Member,~IEEE,}~Yi-Zhe Song,~\IEEEmembership{Senior Member,~IEEE,} and Liang Wang,~\IEEEmembership{Fellow, IEEE}
\thanks{
Peng Xu is with School of Computer Science and Engineering, Nanyang Technological University, Singapore.
Yongye Huang is with Alibaba.
Tongtong Yuan is with Beijing University of Posts and Telecommunications.
Tao Xiang and Yi-Zhe Song are with University of Surrey, United Kingdom.
Timothy M. Hospedales is with University of Edinburgh, United Kingdom.
Liang Wang is with Institute of Automation, Chinese Academy of Sciences, China.
}
\thanks{$^{*}$~Corresponding to Peng Xu, email:~peng.xu@ntu.edu.sg. This work was done before Peng joined NTU.}
}


\maketitle

\begin{abstract}
In this paper, we study learning semantic representations for million-scale free-hand sketches.
This is highly challenging due to the domain-unique traits of sketches, \eg, diverse, sparse, abstract, noisy.
We propose a dual-branch CNN-RNN network architecture to represent sketches, which simultaneously encodes both the static and temporal patterns of sketch strokes.
Based on this architecture, we further explore learning the sketch-oriented semantic representations in two challenging yet practical settings, \ie, hashing retrieval and zero-shot recognition on million-scale sketches.
Specifically, we use our dual-branch architecture as a universal representation framework to design two sketch-specific deep models:
(i) We propose a deep hashing model for sketch retrieval, where a novel hashing loss is specifically designed to accommodate both the abstract and messy traits of sketches. 
(ii) We propose a deep embedding model for sketch zero-shot recognition, via collecting a large-scale edge-map dataset and proposing to extract a set of semantic vectors from edge-maps as the semantic knowledge for sketch zero-shot domain alignment. 
Both deep models are evaluated by comprehensive experiments on million-scale sketches and outperform the state-of-the-art competitors.
\end{abstract}



\IEEEpeerreviewmaketitle

\section{Introduction}
\label{sec:introduction}
With the prevalence of touchscreen devices in recent years, more and more sketches are spreading on the internet, bringing new challenges to the sketch research community.
This has led to a flourishing in sketch-related research~\cite{xu2020deep},
including sketch recognition~\cite{eitz2012humans,xu2019multi},
sketch-based image retrieval (SBIR)~\cite{sangkloy2016sketchy,xu2017cross},
sketch segmentation~\cite{schneider2016sketchSegmentation},
sketch generation~\cite{ha2017sketchrnn,Chen_2018_CVPR}, \etc.
However, sketches are essentially different from natural photos.
In previous work, two unique traits of free-hand sketches had been mostly overlooked:
(i) sketches are highly abstract and iconic, whereas photos are pixel-perfect depictions,
(ii) sketching is a dynamic process other than a mere collection of static pixels.
Exploring the inherent traits of sketch needs the large-scale and diverse dataset of stroke-level free-hand sketches,
since more data samples are required to broadly capture
(i) the substantial variances on visual abstraction, and (ii) the highly complex temporal stroke logic.
However, learning semantic representations for large-scale free-hand sketches is still under-studied.

In this paper, we aim to study learning semantic representations for million-scale free-hand sketches to explore the unique traits of large-scale sketches.
Thus, we use a dataset of $3,829,500$ free-hand sketches,
which is randomly sampling from every category of the Google QuickDraw dataset \cite{ha2017sketchrnn} and termed as ``QuickDraw-3.8M''.
This dataset is highly noisy when compared with TU-Berlin~\cite{eitz2012humans} and Sketchy~\cite{sangkloy2016sketchy},
for that (i) users had only $20$ seconds to draw, and
(ii) no specific post-processing was performed.
In this paper, all our experiments are evaluated on QuickDraw-3.8M.

Since the aforementioned unique traits of sketch make it hard to be represented, we combine RNN stroke modeling with conventional CNN under a dual-branch setting to extract the higher level semantic feature for sketch.
With the RNN stroke modeling branch, dynamic pattern of sketch will be embedded, and by applying CNN branch on the whole sketch, structure pattern of sketch will be encoded.
Accordingly, the spatial and temporal information can be extracted for complete sketch feature learning, based on which,
we further explore learning the sketch-oriented semantic representations in two challenging yet practical settings, \ie, hashing retrieval and zero-shot recognition on million-scale sketches.

Sketch hashing retrieval (SHR) aims to compute an exhaustive similarity-based ranking between a query sketch and all sketches in a very large test gallery.
It is thus a more difficult problem than conventional sketch recognition, since
(i) more discriminative feature representations are needed to accommodate the much larger variations
on style and abstraction, and meanwhile (ii) a compact binary code needs to be learned to facilitate
efficient large-scale retrieval. 
In particular, 
a novel hashing loss that enforces more compact feature clusters for each sketch category in Hamming space is proposed together with the classification and quantization losses. 

{Sketch zero-shot recognition~(SZSR) is more difficult than photo zero-shot learning due to the high-level abstraction of sketch.}
{Its~main challenge is how to choose reliable prior knowledge to conduct
the domain alignment to classify the \textit{unseen} classes.
We propose a sketch-specific deep embedding model using edge-map vector to achieve the domain alignment for SZSR,
whilst all existing zero-shot learning (ZSL) methods designed for photo classification exploit the conventional semantic embedding spaces (\eg, word vector, attribute)
as the bridge for knowledge transfer~\cite{kodirov2015unsupervised}.
To extract high-quality edge-map vector, a large-scale edge-map dataset
(totally $ 290,281$ edge-maps corresponding to $345$ sketch categories) has been collected and released by us.}

The main contributions of this paper can be summarized as follows.
\begin{itemize}
\item {We propose a novel dual-branch CNN-RNN network architecture that simultaneously encodes both the static and temporal patterns of sketch strokes to learn the more fine-grained feature representations.} 
We find that stroke-level temporal information is indeed helpful in sketch feature learning and representing.
\item Based on this architecture, we further explore learning the sketch-oriented semantic representations in two challenging yet practical settings, \ie, hashing retrieval and zero-shot recognition on million-scale sketches.
Specifically, we use our dual-branch architecture as a universal representation framework to design two sketch-specific deep models: 
(a) We propose a deep hashing model for sketch retrieval, where a novel hashing loss is specifically designed to accommodate the abstract nature of sketches, especially on million-scale dataset where the noise becomes more intense. More specifically, we propose a sketch center loss to learn more compact feature clusters for each object category.
(b) We propose a deep embedding model for sketch zero-shot recognition, via collecting a large-scale edge-map dataset and proposing to extract a set of semantic vectors from edge-maps as the semantic knowledge for sketch zero-shot domain alignment. 
The large-scale edge-map dataset that we collect contains 290,281 edge-maps corresponding to 345 sketch categories of QuickDraw.
\end{itemize}

A preliminary conference version of this work has been presented in~\cite{xu2018sketchmate}.
The main extensions can be summarized as: 
(i) 
Based on our dual-branch architecture, we further study learning the sketch-oriented semantic representations in another challenging setting, \ie, zero-shot recognition on million-scale sketches.
We propose a deep embedding model for sketch zero-shot recognition, via collecting a large-scale edge-map dataset and proposing to extract a set of semantic vectors from edge-maps as the semantic knowledge for sketch zero-shot domain alignment. 
(ii) Extensive experiments for sketch zero-shot recognition are conducted. For sketch hashing, more comparisons with the state-of-the-art hashing approaches are also added into this journal version.
(iii) Clearer mathematical formulation and more in-depth analysis are supplemented.

The rest of this paper is organized as follows.
Section~\ref{sec:relatedwork} briefly summarizes related work.
Methodology is provided in Section~\ref{sec:Methodology}:
Experimental results and discussion are presented in Section~\ref{sec:experiments}.
Finally, we draw some conclusions in Section~\ref{sec:conclusion}.

\begin{figure*}[!t]
\begin{center}
\includegraphics[width=\textwidth]{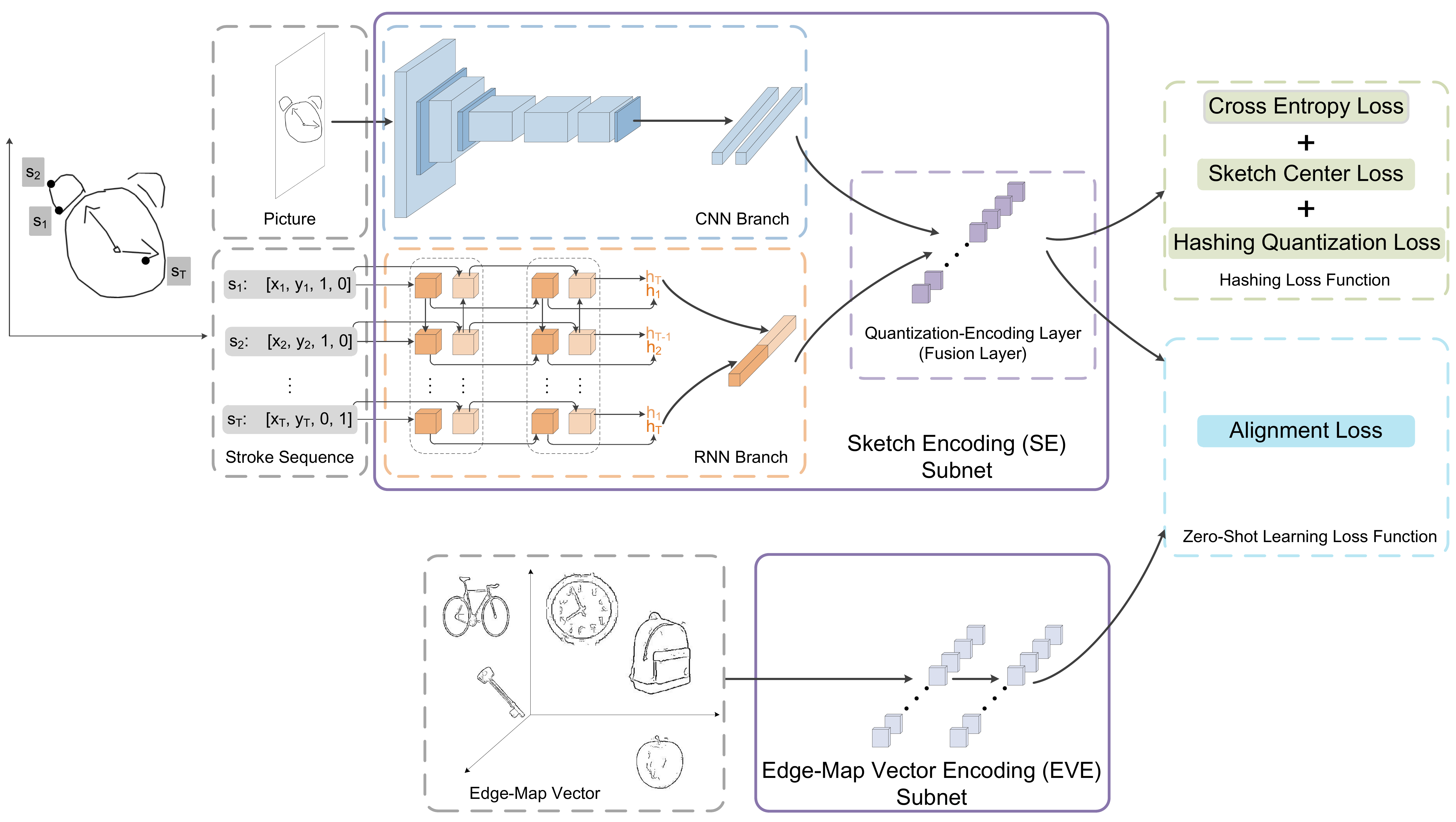}
\end{center}
   \caption{An illustration of our dual-branch CNN-RNN based deep sketch hashing retrieval and zero-shot recognition network. Best viewed in color.}
\label{fig:pipeline}
\end{figure*}

\section{Related Work}
\label{sec:relatedwork}
\subsection{Bottleneck of the Existing Sketch-Related Research}
Sketch research community lacks large-scale free-hand sketch datasets to date,
especially those comparable to the scale of mainstream photo datasets \cite{deng2009imagenet}.
Few medium-scale sketch datasets exist \cite{eitz2012humans, sangkloy2016sketchy}.
They were mainly collected by resorting to crowd-sourcing platforms (\eg, Amazon Mechanical Turk)
and asking the participants to either draw by hand or using a mouse.
Albeit being large enough to train deep neutral networks,
their sizes normally range from hundreds to thousands,
thus inappropriate for large-scale sketch hashing retrieval.
Very recently,
this problem has been alleviated by Ha and Eck \cite{ha2017sketchrnn},
who contributed a large-scale dataset containing $50$ millions of
sketches crossing $345$ categories. These sketches are collected as part
of a drawing game where participants has only $20$ seconds to draw,
hence are often very abstract and noisy. In this paper, we leverage
on this million-scale dataset and study the challenging problems of learning semantic representations for sketch,
while proposing means of tackling the sketch-specific traits of abstraction and temporal ordering.
\subsection{Deep Hashing Learning}
Hashing is an important research topic for fast image retrieval.
Conventional hashing methods~\cite{andoni2006near,weiss2009spectral,gong2013iterative}
mainly utilize hand-crafted features as image representations
and propose various projections and quantization strategies to learn the hashing codes.
Recently, deep hashing learning has shown superiority on better preserving the semantic information
when compared with shallow methods~\cite{xia2014supervised,lin2015deep,shen2015supervised}.
In the initial attempt, feature representation and hashing codes were learned in separate stages~\cite{xia2014supervised},
where subsequent work ~\cite{lin2015deep,zhao2015deepsemantic,liu2016deepsupervised}
suggested superior practice through joint end-to-end training.
To our best knowledge,
only few previous works~\cite{liu2017deepsketchhashing, shen2018zero}
have specifically designed the deep hashing frameworks targeted on sketch data to date.
Despite their superior performances,
sketch specific traits such as stroke ordering and drawing abstraction were not accommodated for.
The dataset \cite{sangkloy2016sketchy} that they evaluated on is also arguably too small to truly show
for the practical value of a deep hashing framework.
We address these issues by working with a much larger free-hand sketch dataset,
and designing sketch-specific solutions that are crucial for million-scale sketch retrieval.

\subsection{Zero-Shot Learning}

Most existing zero-shot learning~(ZSL) methods in computer vision community are studied on photos and videos, which use semantic spaces as the bridge for knowledge transfer~\cite{kodirov2015unsupervised} by assuming the \textit{seen} and the \textit{unseen} classes share a common semantic
space.
The semantic spaces can be defined by word vector~\cite{socher2013zero,lizhang2017zero}, sentence description~\cite{reed2016learning},
and attribute vector~\cite{farhadi2009describing,parikh2011relative}.
Moreover, many existing zero-shot learning models are embedding models.
According to different embedding spaces used, these embedding models can be categorized as
three groups: (i) embed visual features into semantic space~\cite{socher2013zero,frome2013devise,fu2016semi,song2018transductive},
(ii)~embed semantic vectors into visual space~\cite{lizhang2017zero}, and
(iii)~embed visual features and semantic vectors into a common intermediate
space~\cite{lei2015predicting,romera2015embarrassingly}.
Recently, several methods~\cite{shen2018zero,yelamarthi2018zero}
are proposed on free-hand sketch based image retrieval~(SBIR), leaving the more intrinsically~theoretical problem
sketch zero-shot recognition under-studied.
In this paper, we argue that,
compared with other data modalities, edge-map has smaller domain gap to sketch domain.
Considering this superiority of edge-map domain,
we propose to use the semantic information extracted from edge-map domain as knowledge to guide the SZSR domain alignment.
This is different from the technique for photo ZSL that conducts the domain alignment based on conventional semantic knowledge (\eg, word vector, sentence description).

\section{Methodology}
\label{sec:Methodology}

\subsection{Sketch Representation by a Dual-Branch CNN-RNN Architecture}
\label{sec:Rep}
Let $\mathcal{K} = \{{K}_n=({\bf P}_n,{\bf S}_n)\}_{n=1}^{N}$ be $N$ sketch~samples crossing $L$ possible categories
and $\mathcal{Y} = \{y_n\}_{n=1}^{N}$ be their respective category labels.
Each sketch sample ${K}_n$ consists of a sketch picture ${\bf P}_n$ in raster pixel space and a corresponding sketch stroke sequence ${\bf S}_n$. 
We aim to learn the deep sketch representation, which can better handle the domain-unique traits of free-hand sketches and benefit to various sketch-oriented tasks.

As aforementioned analysis, learning discriminative sketch representations is a very challenging task
due to the high degree of variations in style and abstraction.
This problem is made worse under a large-scale sketch based setting since better feature representations are needed for more fine-grained feature comparison.
Despite shown to be successful on a much smaller sketch dataset \cite{yu2017sketch},
CNN-based network completely abandons the natural point-level and stroke-level temporal information of free-hand sketches,
which can now be modeled by a RNN network, thanking to the seminal work by \cite{ha2017sketchrnn}.
In this paper, we for the first time propose to combine the best from the both worlds for free-hand sketches
-- utilizing CNN to extract abstract visual concepts and RNN to model human sketching temporal orders.

As shown in Figure \ref{fig:pipeline}, the RNN branch adopts bidirectional Gated Recurrent Units for the stroke-level temporal information extraction, whose output is the concatenation of their last hidden layers. The CNN branch can apply any kinds of architectures designed for photos, \eg, AlexNet, ResNet, with the last fully connected layer as the output. It should be noted that several 
sketch-specific data preprocessing operations
are adopted, which will be illustrated clearly in the experiment part. 
Finally, we conduct branch interaction via a late-fusion layer (one fully connected layer with specific activation)
by concatenation.
This late-fusion layer will provide representations for various tasks.
\cut{
\todo ?
Through additional discriminative power offered by the tasks injected in, we hope this can lead to better feature learning.}

\subsection{Deep Hashing for Large-Scale Sketch Retrieval}
\label{sec:HASH}
\subsubsection{Problem Formulation}
With the provided data $\mathcal{K} = \{{K}_n=({\bf P}_n,{\bf S}_n)\}_{n=1}^{N}$, we aim to learn a mapping $\mathcal{M}:\mathcal{K} \rightarrow \{0, 1\}^{D \times N}$,
which represents sketches as $D$-bit binary codes
${\bf B} = \{{\bf b}_n\}_{n=1}^{N} \in \{0, 1\}^{D \times N}$, while maintaining relevancy in accordance with the semantic and visual similarity amongst sketch data.
To achieve this, we transform the CNN-RNN late fusion layer as the quantization-encoding layer 
by
proposing a novel objective with specifically designed losses, which will be elaborated in the following subsections.

\begin{figure*}[!t]
	\centering
	\subfigure[cross entropy loss]{
		\label{fig:softmax_fig}
		\includegraphics[width=0.3\textwidth]{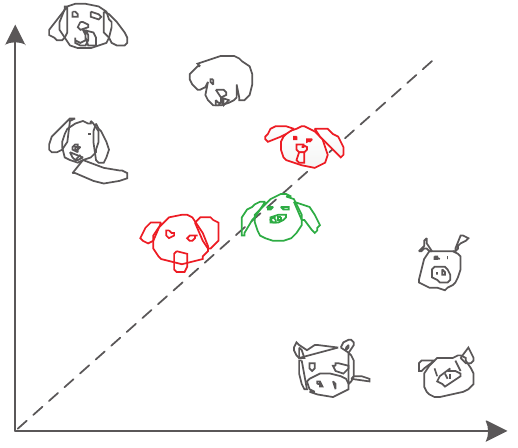}}
	\subfigure[cross entropy loss + common center loss]{
		\label{fig:softmax_centerloss_fig}
		\includegraphics[width=0.3\textwidth]{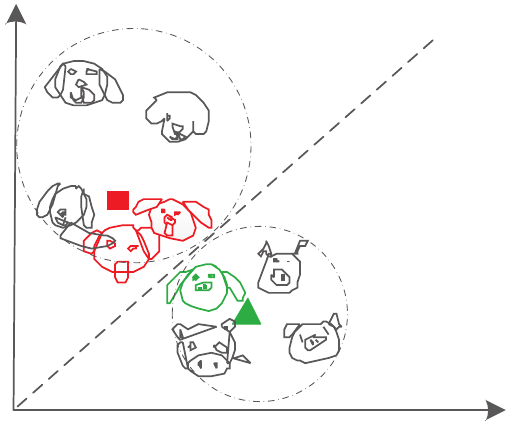}}
    \subfigure[cross entropy loss + sketch center loss]{
		\label{fig:softmax_sketchcenterloss_fig}
		\includegraphics[width=0.3\textwidth]{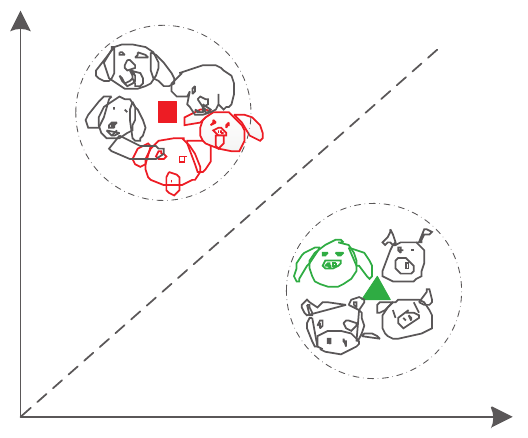}}
	\caption{Geometric interpretation of sketch feature layouts obtained by different loss functions. The dashed line denotes the softmax decision boundary. See details in text.}
	\label{fig:softmax_centerloss}
\end{figure*}

\subsubsection{Classification-Oriented Pretraining}
\label{sec:CNN-RNN}
Due to the intrinsic difference between CNN and RNN, we adopt staged-training to construct our sketch-specific dual-branch CNN-RNN (CR) network. In first stage, we individually pretrain CNN and RNN branches using cross entropy loss, while CNN branch takes in the raster pixel sketch and RNN branch takes in the corresponding stroke sequence vector. In second stage, we conduct branch interaction via a late-fusion layer (one fully-connected layer with sigmoid activation) by concatenation that the amount of neurons in fusion layer equals to the hashing code length set by us. In later stages, this late-fusion layer will be transformed as quantization-encoding layer after we add the binary constraints. For the pretraining and fusion in these two stages, we use cross entropy loss (CEL) as our loss function since for million-scale dataset
 (i) compared with category labels, other more detailed annotations (\eg,
pairwise label~\cite{liu2016deepsupervised}, triplet label~\cite{wangxiaofang2016deep}) are 
expensive, and (ii) the pairwise or triplet contrastive training with high runtime complexity is also unrealistic.

In this paper, cross entropy loss (CEL) 
goes as
\begin{equation}
\label{equ:cel_loss}
\begin{split}
\mathcal{L}_{cel} = & \frac{1}{N} \sum_{n = 1}^{N} - \log \frac{\mathrm{e}^{{\bf W}_{y_n}^{T}{\bf f}_n+\widehat{b}_{y_n}}}{\sum_{j=1}^{L} \mathrm{e}^{{{\bf W}_j^{T}{\bf f}_n+\widehat{b}_j}}},
\end{split}
\end{equation}
where ${\bf W}_j \in \mathbb{R}^{D}$ is the $j$th column of the weights ${\bf W} \in \mathbb{R}^{D \times L}$ between the late-fusion layer and $L$-way softmax outputs. ${\bf f}_{n}$ is the low-dimensional real-valued feature that will play the role of hashing feature in our later hashing stages.~$\widehat{b}_j$ is the $j$th term of the bias $\widehat{{\bf b}} \in \mathbb{R}^L$.


\subsubsection{Sketch Center Loss}
\label{sec: scl}
In theory, cross entropy loss can perform reasonably well on discriminating category-level semantics, however,
our used large-scale sketch dataset presents an unique challenge -- sketch are highly abstract,
often making semantically different categories to exhibit similar appearance (see Figure \ref{fig:softmax_centerloss}(a)
 for an example of `dog' vs. `pig'). We need to make sure such abstract nature of sketches do not hinder overall retrieval performance.

The common center loss (CL) was proposed in \cite{wen2016discriminative} to tackle such a problem by introducing the
concept of class center,
${\bf c}^{y_{n}}$, to characterize the intra-class variations.
Class centers should be updated as deep features change, in other words,
the entire training set should be taken into account and features of every class should be averaged in each iteration.
This is clearly unrealistic and normally compromised by updating only within each mini-batch.
This problem is even more salient under our sketch hashing retrieval setting -- (1)
for million-scale hashing, updating common center within each mini-batch can be highly inaccurate and
even misleading (as shown in later experiments), and this problem is worsened by the abstract nature of sketches
in that only seeing sketches within one training batch doesn't necessarily provide
useful and representative gradients for class centers; (2) despite of more compact internal category structures
(Figure \ref{fig:softmax_centerloss}(b))
 with common center loss, there is no explicit constraint to set apart between each, as a direct comparison with
 Figure \ref{fig:softmax_centerloss}(c).

These issues call for a sketch-specific center loss that is able to deal with million-scale hashing retrieval.
For sketch hashing, we need compact and discriminative features to aggregate samples belonging
to the same category and segregate the visually confusing categories.
Thus, a natural intuition would be: it is possible if we can find a \textit{fixed} but \textit{representative} center feature
for each class, to avoid the computational complexity during training, and
meanwhile enforcing semantic separation between sketch categories.

We propose \textit{sketch center loss} that is specifically designed for million-scale sketch hashing retrieval as
\begin{equation}
\label{equ:sketch_center_loss}
\mathcal{L}_{scl} = \frac{1}{N} \sum_{n = 1}^{N} \|{\bf f}_n - {\bf c}^{y_n}\|_2^2~,
\end{equation}
in which $N$ is the total number of training samples.
${\bf f}_n$ is the real-valued hashing feature for $n$th training sample $K_n$ ($n \in [1, N]$)
obtained by late-fusion layer. ${\bf c}^{y_n}$ denotes the feature center value of class $y_n$ ($y_n \in \{1, 2, \ldots, L\}$).
Here, for a sample class $y \in \{1, 2, \ldots, L\}$, the feature center ${\bf c}^{y}$ is a fixed value calculated via
\begin{equation}
\label{equ:sketch_center}
{\bf c}^{y} = \frac{1}{\gamma|\mathcal{K}^y|} \sum_{K_n \in \mathcal{K}^y} \mathbbm{1}(h_{lower}^y < h_n < h_{upper}^y) \widetilde{{\bf f}}_n~,
\end{equation}
in which $\widetilde{{\bf f}}_n$ is the real-valued feature extracted from the late-fusion layer of the second stage pretrained model as illustrated in Section~\ref{sec:CNN-RNN}.
$\mathcal{K}^y$ is the sample set of class $y$ with total number $|\mathcal{K}^y|$.
$h_n$
denotes the image entropy of $n$th sample.
Let $\mathcal{H}^y$ be the image entropy value set of $\mathcal{K}^y$.
If $K_n \in \mathcal{K}^y$, we have $h_n \in \mathcal{H}^y$.
$h_{lower}^y$ and $h_{upper}^y$ are the $\phi$th and $\varphi$th ($\gamma = \varphi \% - \phi \%$ and $0 \leq \phi \% < \varphi \% \leq 1$) percentiles of $\mathcal{H}^y$, respectively.
We define image entropy for sketch data as
\begin{equation}
\label{equ:image_entropy}
H = \sum_{i = 0, 255} - p_i \log p_i~,
\end{equation}
where $p_i$ is the proportion of the gray pixel values $i$ in each sketch.
For sketch, image entropy value is robust to displacement, directional change, and rotation. Stroke shake changes the locus of the drawing lines and will changes the entropy value.
Now, we can jointly use $\mathcal{L}_{cel} + \mathcal{L}_{scl}$ to conduct
the third-stage training for more discriminative feature learning.

Key ingredient to a successful sketch center loss is the guarantee of non-noisy data (outliers),
as it will significantly affect the class feature centers. However, datasets collected with crowdsourcing
without manual supervision are inevitable to noise. Here we propose this noisy data removal technique to greatly alleviate
such issues by resorting to image entropy.

\begin{figure}[!t]
\begin{center}
\includegraphics[width=0.5\textwidth]{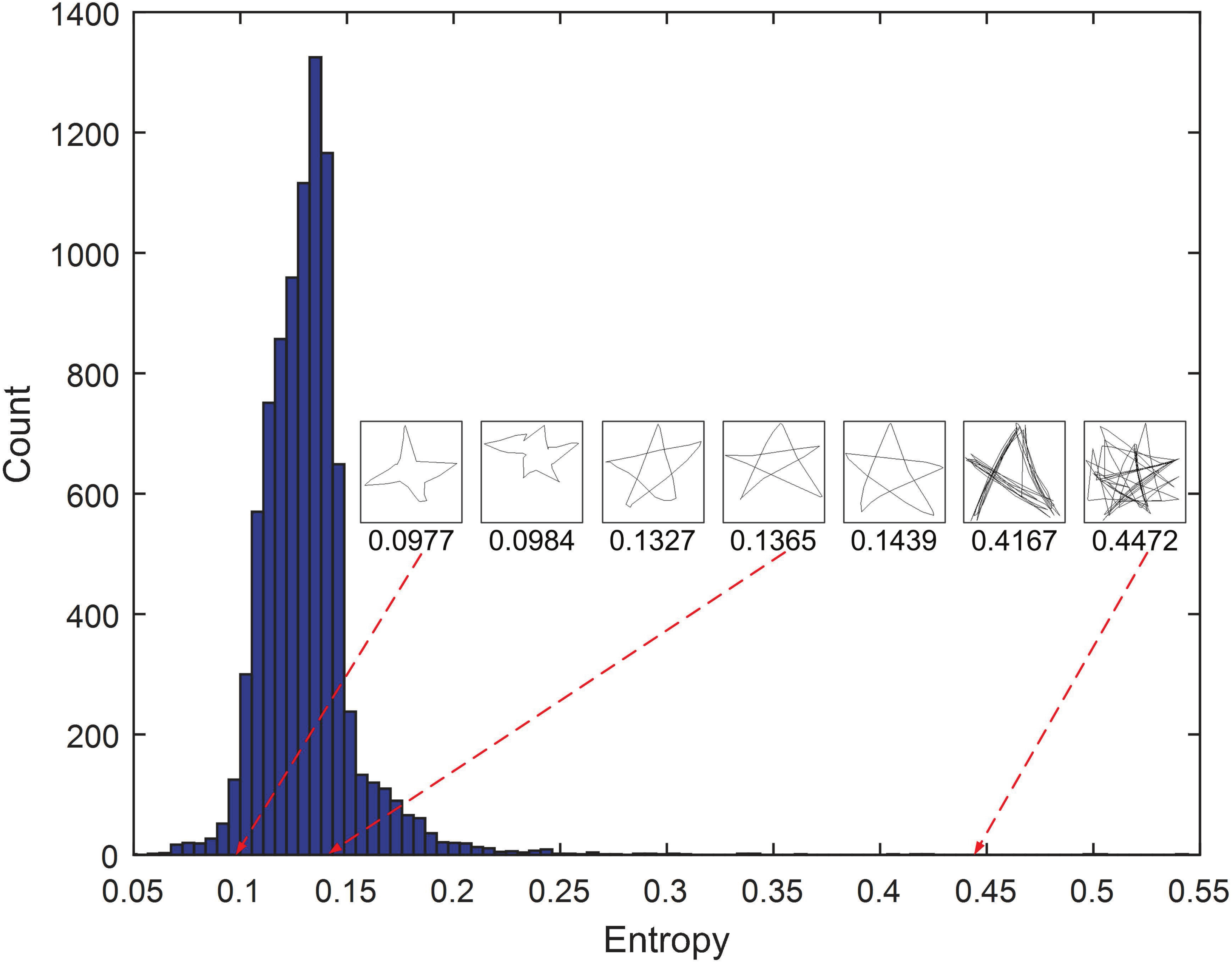}
\end{center}
   \caption{\textcolor{black}{Image entropy histogram of `stars' in our training set.
   The blue bars denote the bin counts within different entropy ranges.
    Some representative sketches corresponding to different entropy values are illustrated. See details in text.}}
\label{fig:star}
\end{figure}

Given a category of sketch, we can get entropy for each sketch and the overall entropy distribution on a category basis.
We empirically find that keeping the middle $90\%$ of each category as normal samples gives us good 
results.
This means that we set $\phi \%$ and $\varphi \%$ 
as $0.05$
and $0.95$, respectively. In Figure~\ref{fig:star}
, we visualize the entropy histogram of ``star'' samples
in our hashing training set~(our data splits can be seen in Table~\ref{table:quickdrawdataset}).
If we choose the middle $90\%$ samples as normal samples for ``star'' category,
we can calculate and obtain the $0.05$ and $0.95$ percentiles of ``star'' entropy are $0.1051$ and $0.1721$, respectively.
Remaining samples~(their entropy values $\in~[0,0.1051)~\bigcup~(0.1721,1]$) can be treated as outliers or noise points.
In Figure~\ref{fig:star}, we can see that low entropy sketches tend to be more abstract,
yet high entropy ones tend to be more messy, while the middle range ones depict good looking stars.

\subsubsection{Quantization and Encoding}
The late-fusion layer can generate low dimensional real-value feature ${\bf f}_{n}$, which will be further transformed to the hashing code ${\bf b}_{n}$. The transformation function goes as follows:
\begin{equation}
\label{equ:quantization}
{\bf b}_n = (sgn({\bf f}_n - {\bf 0.5}) + {\bf 1}) / 2,~~n \in [1, N].
\end{equation}
Therefore, our late-fusion layer also can be termed as quantization-encoding layer as shown in Figure~\ref{fig:pipeline}.
The quantization loss (QL) is used to reduce the error caused by quantization-encoding:
\begin{equation}
\label{equ:ql_loss}
\begin{split}
\mathcal{L}_{ql} = \frac{1}{N} \sum_{n = 1}^{N} \|{\bf b}_n - {\bf f}_n\|_2^2,~ s.t.~{\bf b}_n \in \{0, 1\}^{D}.
\end{split}
\end{equation}

\subsubsection{Full Loss Function}
By combining the above, our full loss function becomes as
\begin{equation}
\label{equ:full}
\mathcal{L}_{full} = \mathcal{L}_{cel} + \lambda_{scl}\mathcal{L}_{scl} + \lambda_{ql}\mathcal{L}_{ql},~ s.t.~{\bf b}_n \in \{0, 1\}^{D},
\end{equation}
where $\lambda_{scl}$~and~$\lambda_{ql}$ control the relative importance of each loss.
We designed a staged-training and alternative iteration strategy to
minimize this binary-constraint loss function.
The detailed training and optimization are described in Algorithm~\ref{alg:1}.

\begin{algorithm}[!t]
    	\caption{The learning algorithm for our proposed deep sketch hashing model.}
  	\label{alg:1}
        \begin{algorithmic}
        \Require $\mathcal{K} = \{{K}_n=({\bf P}_n,{\bf S}_n)\}_{n=1}^{N}$, $\mathcal{Y} = \{y_n\}_{n=1}^{N}$.
        \State 1. Train CNN from scratch using $\{{\bf P}_n\}_{n=1}^{N}, \mathcal{L}_{cel}$.
        \State 2. Train RNN from scratch using $\{{\bf S}_n\}_{n=1}^{N}, \mathcal{L}_{cel}$.
        \State 3. Parallelly connect pretrained CNN and RNN branches via late-fusion layer. Fine-tune the fused model using $\mathcal{L}_{cel}$.
        \State 4. Calculate class feature centers basing on Equation~(\ref{equ:sketch_center}) and the pretrained model in step $3$. Fine-tune the whole network using $\mathcal{L}_{cel} + \lambda_{scl}\mathcal{L}_{scl}$.
        \State 5. Train as the following iterations.~$t$ represents current iteration.
        \For{number of training iterations}
            \For{a fixed number of training epochs}
                \State 6. Fix ${\bf b}_n^{t}$, update $\Theta$ using Equation~(\ref{equ:full}).
            \EndFor
            \State 7. Fix $\Theta$, calculate ${\bf b}_n^{t+1}$ using Equation~(\ref{equ:quantization}).
        \EndFor
        \Ensure Network parameters: $\Theta$. Binary hash code matrix ${\bf B} \in \mathbb{R}^{D \times N}.$
        \end{algorithmic}
    \end{algorithm}

\subsection{Zero-Shot Recognition for Large-Scale Sketch}
\label{sec:ZSL}
\subsubsection{Problem Formulation}
Let $\mathcal{K}_{tr} = \{{K}_i=({\bf P}_i,{\bf S}_i, {\bf v}^{y_i})\}_{i=1}^{M}$ be $M$~labelled~sketch
training samples crossing $L_{tr}$ categories
and $\mathcal{Y}_{tr}$ be their associated category label set
that $y_i \in \mathcal{Y}_{tr}$ and $|\mathcal{Y}_{tr}| = L_{tr}$.
Each sketch training sample ${K}_i$ consists of a sketch ${\bf P}_i$ in raster pixel space and a corresponding
sketch stroke sequence ${\bf S}_i$, and ${\bf v}^{y_i} \in \mathbb{R}^{D_e}$
denotes the associated \textcolor{black}{edge-map} vector of class $y_i$.

Similarly, let $\mathcal{K}_{te} = \{{K}_j=({\bf P}_j,{\bf S}_j)\}_{j=1}^{N-M}$ be $N-M$~sketch
testing samples crossing $L_{te}$ possible categories.
$\mathcal{Y}_{te}$ is the possible category label set of $\mathcal{K}_{te}$, and $|\mathcal{Y}_{te}| = L_{te}$.

Given a test sketch ${K}_j$, the goal of zero-shot recognition is to predict a class label $y_j \in \mathcal{Y}_{te}$.
We have $\mathcal{Y}_{tr} \bigcap \mathcal{Y}_{te} = \emptyset$, \ie,
the training (\textit{seen}) classes and test (\textit{unseen})~classes are disjoint.
Each \textit{seen} or \textit{unseen} class is associated with a pre-defined edge-map based semantic vector
${\bf v}^{y_i}$ or ${\bf v}^{{y}_j}$
, referred to the edge-map based visual class prototype.
Please see how to define and obtain edge-map vectors in following sections.

\subsubsection{Sketch-Specific Zero-Shot Recognition Model}
\label{sec:ourZSLmodel}

%
As shown in Figure~\ref{fig:pipeline}, 
our sketch zero-shot recognition pipeline involves two subnets: (i) sketch encoding~(SE) subnet,
(ii) edge-map vector
encoding~(EVE) subnet.

The sketch encoding subnet is implemented by our proposed sketch-specific dual-branch CNN-RNN network,
which takes ${\bf P}_i$~and~${\bf S}_i$ as input and outputs a feature vector
${\mathcal{F}}_{\Theta_{S}} ({\bf P}_i, {\bf S}_i) \in \mathbb{R}^{D_s}$.
${\mathcal{F}}_{\Theta_{S}}$ denotes feature extraction by the SE subnet, which is parameterised by $\Theta_{S}$.
The edge-map vector encoding subnet is implemented by two fully-connected layers, which takes in ${\bf v}^{y_i}$
and outputs an embedding vector
${\mathcal{F}}_{\Theta_{E}} ({\bf v}^{y_i}) \in \mathbb{R}^{D_{s}}$.

The loss function of our zero-shot model is
\begin{equation}
\label{equ:zeroshot_loss}
\begin{split}
\mathcal{L}_{z} = \frac{1}{M} \sum_{i = 1}^{M} \| {\mathcal{F}}_{\Theta_{S}} ({\bf P}_i, {\bf S}_i) - {\mathcal{F}}_{\Theta_{E}} ({\bf v}^{y_i})\|_2^2 + \lambda \| \Theta_{E}\|_2^2,
\end{split}
\end{equation}
where $\lambda$ is the weighting coefficient.

In testing, given the test sketch~${K}_j$, its class label prediction is its nearest embedded edge-map visual class prototype
in the sketch space,
\begin{equation}
\label{equ:zeroshot_testing}
\begin{split}
y_{j} = \mathop{\arg\min}_{\hat{y}} \| {\mathcal{F}}_{\Theta_{S}} ({\bf P}_j, {\bf S}_j) - {\mathcal{F}}_{\Theta_{E}} ({\bf v}^{\hat{y}})\|_{2}^{2},
\end{split}
\end{equation}
where ${\bf v}^{\hat{y}}$ is the possible class prototype.
Our detailed training and optimization are described in Algorithm~\ref{alg:2}.

\begin{algorithm}[!t]
    	\caption{The learning algorithm for our proposed deep embedding model for sketch zero-shot recognition.}
  	\label{alg:2}
        \begin{algorithmic}
        \Require $\mathcal{K}_{tr} = \{{K}_i=({\bf P}_i,{\bf S}_i, {\bf v}^{y_i})\}_{i=1}^{M}$.
        \State 1. Train our SE subnet in following stages:
        \State (1.1) Train CNN from scratch using $\{{\bf P}_i\}_{i=1}^{M}, \mathcal{L}_{cel}$.
        \State (1.2) Train RNN from scratch using $\{{\bf S}_i\}_{i=1}^{M}, \mathcal{L}_{cel}$.
        \State (1.3) Parallelly connect pretrained CNN and RNN branches via late-fusion layer. Fine-tune the fused model using $\mathcal{L}_{cel}$ and obtain $\Theta_{S}$.
        \State 2. Fix $\Theta_{S}$, train our EVE subnet using Equation~(\ref{equ:zeroshot_loss}).

        \Ensure Network parameters: $\Theta_{S}$ and $\Theta_{E}$.
        \end{algorithmic}
    \end{algorithm}

\subsection{Large-Scale Edge-Map Dataset}
\label{sec:our-edgemap-dataset}
We need a high-quality edge-map dataset to obtain reliable edge-map vectors.
Considering that we want to use edge-map vectors as knowledge to achieve the domain alignment for SZSR,
the edge-map dataset that we need has to cover all the sketch categories we used.
Therefore, we collect and contribute the first large-scale edge-map dataset, which contains $290, 281$ edge-maps
corresponding to $345$ sketch categories of QuickDraw~\cite{ha2017sketchrnn}.
In this section, we describe the two steps of our data collection: photo collection and edge-map extraction.
\begin{figure*}[!t]
\begin{center}
\includegraphics[width=\textwidth]{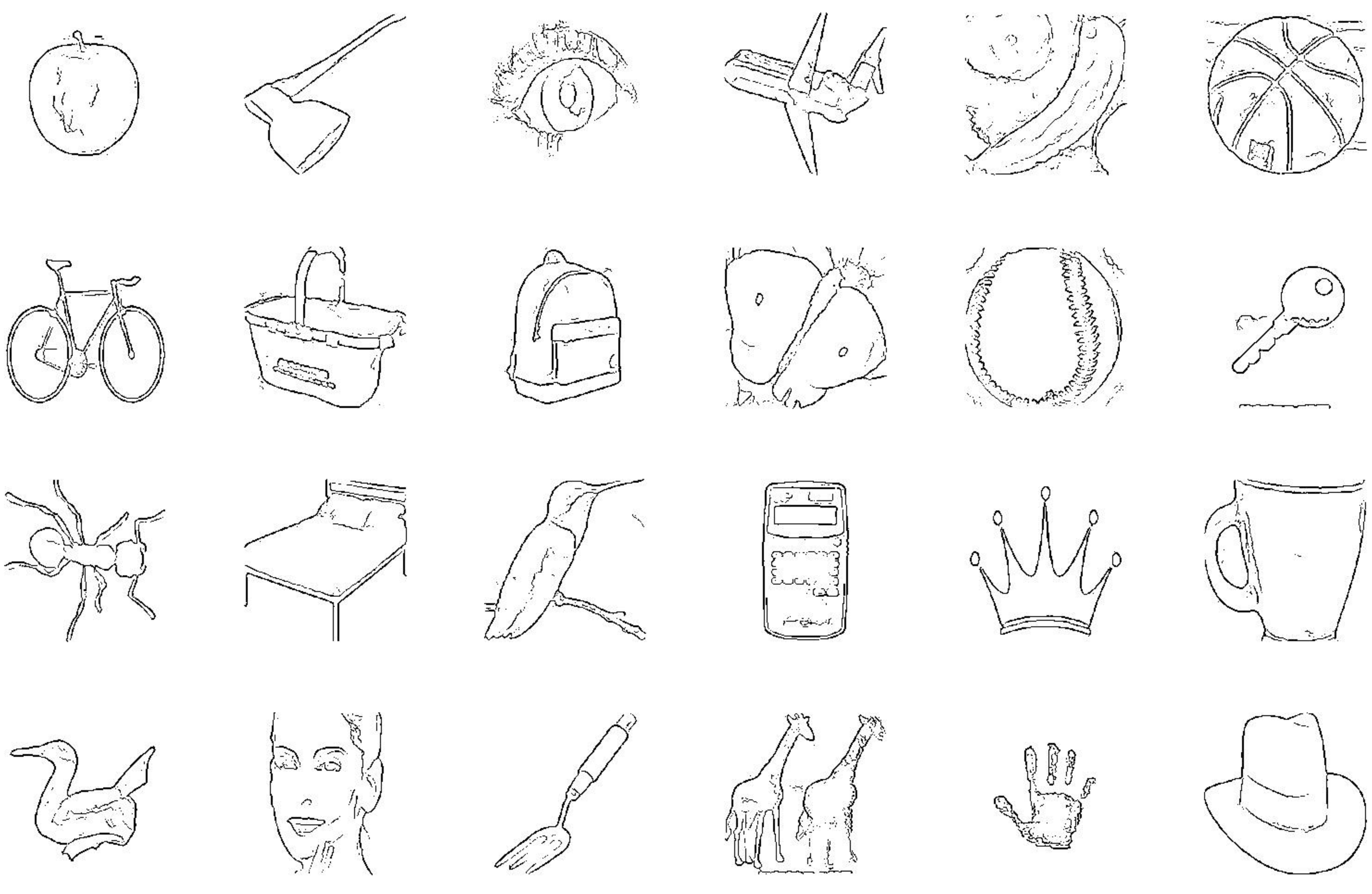}
\end{center}
\caption{Samples~(apple, axe, eye, airplane, banana, basketball, bicycle, basket,
   backpack, butterfly, baseball, key, ant, bed, bird, calculator, crown, cup, duck, face, fork, giraffe, hand, hat) of our collected edge-map dataset.}
\label{fig:edgemap_samples}
\end{figure*}
\subsubsection{Photo Collection}
The first step to collect edge-maps is collecting photos corresponding to the sketch categories.
As earlier discussion in Section~\ref{sec:ZSL}, we aim to utilize edge-maps to conduct sketch-specific
zero-shot learning on QuickDraw dataset~\cite{ha2017sketchrnn}. Thus, our edge-map categories need to cover all
the~$345$ (\textit{seen} and \textit{unseen}) sketch categories.
There are $157$~(out of $345$) classes of QuickDraw having at least one corresponding class in ImageNet~\cite{ILSVRC15}.
For each of these $157$ categories, we manually choose one most similar category from ImageNet.
Since many of ImageNet are multi-object images,
we crop photos from the annotated bounding box areas, thus only photos with bounding box annotations provided can be used.
This cropping operation can alleviate the domain gap between edge-map and sketch.
For the remaining $188$~categories, we program crawler to download images from Google Images
\footnote{\url{https://images.google.com}}.

\subsubsection{Edge-Map Extraction}
After photo collection, we use {Edge Boxes toolbox}~\cite{zitnick2014edge} to extract edge-maps from our collected photos.
Each edge-map has been located in the center and takes up about $90\%$ area of the photo.
We resized all the edge-maps as $3 \times 224 \times 224$.
We recruit five volunteers to manually remove the messy edge-maps that can not be recognize by human.
Finally, we~obtain~$290, 281$ edge-maps across $345$ categories (averagely $841$ per category).
Some samples of our edge-map dataset are shown in Figure~\ref{fig:edgemap_samples}.
Our edge-map dataset is available at
\url{https://github.com/PengBoXiangShang/EdgeMap345C_Dataset}.

\begin{table}[!t]
\small
\begin{center}
\caption{Dataset splits of QuickDraw-3.8M.}
\label{table:quickdrawdataset}
\resizebox{\columnwidth}{!}{
\begin{tabular}{p{1.5cm}<{\centering}|| c | p{3.5cm}<{\centering}}
\hline
 Splits & Number per Category & Amount \\
\hline
\hline
 Training & 9000 & $9000\times345=3105000$ \\
 Validation & 1000 & $1000\times345=345000$ \\
 Gallery & 1000 & $1000\times345=345000$ \\
 Query & 100 & $100\times345=34500$ \\
\hline
\end{tabular}}
\end{center}


\end{table}

\begin{table*}[!t]
\footnotesize
\begin{center}
\caption{Comparison with the state-of-the-art deep hashing methods and our model variants on QuickDraw-3.8M retrieval gallery.}
\label{table:compare_with_deep_baseline}
\resizebox{\textwidth}{!}{
\begin{tabular}{l||c|c|c|c||c|c|c|c}
\hline
 \multirow{2}{*}{Model} & \multicolumn{4}{c||}{Mean Average Precision} & \multicolumn{4}{c}{Precision @200} \\
\cline{2-9}
 & 16 bits & 24 bits & 32 bits & 64 bits & 16 bits & 24 bits & 32 bits & 64 bits \\
\hline\hline
 DLBHC~\cite{lin2015deep} & 0.5453 & 0.5910 & 0.6109 & 0.6241 & 0.5142 &0.5917 &0.6169 &0.6403 \\
 DSH-Supervised~\cite{liu2016deepsupervised,liuhaomiao2019ijcv} & 0.0512 & 0.0498 & 0.0501 & 0.0531 & 0.0510 &0.0512 &0.0501 &0.0454  \\
 DSH-Sketch~\cite{liu2017deepsketchhashing} & 0.3855 & 0.4459 & 0.4935 & 0.6065 & 0.3486 &0.4329 &0.4823 &0.6040 \\
 GreedyHash ~\cite{su2018greedy} & 0.4127  & 0.4520  &  0.4911  &  0.5816   &  0.3896  & 0.4511 & 0.4733  & 0.5901 \\
\hline
\hline
 CR+CEL & 0.5969 &0.6196 & 0.6412 & 0.6525 & 0.5817 & 0.6292 &0.6524 &0.6730     \\
 CR+CEL+CL & 0.5567 & 0.5856 & 0.5911 & 0.6136 &  0.5578 & 0.6038 &0.6140 &0.6412      \\
 CR+CEL+SCL & 0.6016 & 0.6371 & 0.6473 & 0.6767 &  0.5928 & 0.6298 & 0.6543&0.6875   \\
\hline
\hline
 CR+CEL+SCL+QL (Our Full Hashing Model) & \textbf{0.6064} & \textbf{0.6388} &\textbf{0.6521} & \textbf{0.6791} &  \textbf{0.5978} & \textbf{0.6324} &\textbf{0.6603} &\textbf{0.6882}   \\
\hline
\end{tabular}
}
\end{center}

\end{table*}

\begin{table*}[!t]
\footnotesize
\begin{center}
\caption{Comparison with shallow hashing competitors on QuickDraw-3.8M retrieval gallery.}
\label{table:shallow_hash_method_results}
\resizebox{\textwidth}{!}{
\begin{tabular}{c | c||c|c|c|c|c|c||c|c}
\hline
 \multicolumn{2}{c||}{~} & \multicolumn{6}{c||}{Unsupervised} &  \multicolumn{2}{c}{Supervised} \\
 \cline{3-10}
\multicolumn{2}{c||}{~} & PCA-ITQ~\cite{gong2013iterative} & LSH~\cite{andoni2006near} & SH~\cite{weiss2009spectral} & SKLSH~\cite{raginsky2009locality} & DSH~\cite{jin2014density} & PCAH~\cite{wang2006PCAH} &  SDH~\cite{shen2015supervised} & CCA-ITQ~\cite{gong2013iterative} \\
\hline\hline
 \multirow{4}{*}{HOG} & 16 bits & 0.0222 & 0.0110 & 0.0166 & 0.0096 &	0.0186 & 0.0166 & 0.0160 & 0.0185 \\
                      & 24 bits & 0.0237 & 0.0121 & 0.0161 & 0.0105 &	0.0183 & 0.0161 & 0.0186 & 0.0195 \\
                      & 32 bits & 0.0254 & 0.0128 & 0.0156 & 0.0108 &	0.0224 & 0.0155 & 0.0219 & 0.0208 \\
                      & 64 bits & 0.0266 & 0.0167 & 0.0157 & 0.0127 & 0.0243 & 0.0146 & 0.0282 & 0.0239 \\
\hline
\hline
 \multirow{4}{*}{deep feature} & 16 bits & 0.4414 & 0.3327 & 0.4177 & 0.0148 & 0.3451 & 0.4375 & 0.5781 & 0.3638 \\
                      & 24 bits & 0.5301 & 0.4472 & 0.5102 & 0.0287 & 0.4359 & 0.5224 & 0.6045 & 0.4623 \\
                      & 32 bits & 0.5655 & 0.5001 & 0.5501 & 0.0351 & 0.4906 & 0.5576 & 0.6133 & 0.5168 \\
                      & 64 bits& \textbf{0.6148} & \textbf{0.5801} & \textbf{0.5956} & \textbf{0.0605} & \textbf{0.5718} & \textbf{0.6056} & \textbf{0.6273} & \textbf{0.5954} \\
\hline
\end{tabular}
}
\end{center}

\end{table*}

\section{Experiments}
\label{sec:experiments}
\subsection{Hashing}
\subsubsection{Datasets and Settings}
Google QuickDraw dataset \cite{ha2017sketchrnn}
contains 345 object categories with more than 100,000 free-hand
sketches for each category. Despite the large-scale sketches publicly available,
we empirically find out that a number of around 10,000 sketches suffices for a sufficient
representation of each category and thus randomly choose  9000, 1000 from which for training and validation, respectively.
For evaluation, we form our query and retrieval gallery set by randomly choosing 100 and 1000 sketches from each category.
A detailed illustration of the dataset split can be found in Table~\ref{table:quickdrawdataset}.
Overall, this constitutes an experimental dataset of 3,829,500 sketches, standing itself on a million-scale analysis
of sketch specific hashing problem, an order of magnitude larger than previous state-of-the-art
research \cite{liu2017deepsketchhashing}, which we term as ``QuickDraw-3.8M".
We scale the raster pixel sketch to $3 \times 224 \times 224$, with each brightness channel tiled equally,
while processing the vector sketch same as with \cite{ha2017sketchrnn},
with one critical exception -- rather than treating pen state as a sequence of three binary switches,
\ie, continue ongoing stroke, start a new stroke and stop sketching,
we reduce to two states by eliminating the sketch termination signal for faster training,
leading each time-step input to the RNN module a four-dimensional input.

\subsubsection{Implementation Details}
Our RNN-based encoder uses bidirectional Gated Recurrent Units with two layers,
with a hidden size of $512$ for each layer,
and the CNN-based encoder follows the AlextNet \cite{krizhevsky2012imagenet} architecture
with major difference at removing the local response normalization for faster training.
We implement our model on one single Pascal TITAN X GPU card, where for each pretraining stage,
we train for $20, 5, 5$ epochs, taking about $20, 10, 10$ hours respectively.
We set the importance weights $\lambda_{scl} = 0.01$ and $\lambda_{ql} = 0.0001$ during training
and find this simple strategy works well. The model is trained end to end using the
Adam optimizer \cite{kingma2014adam}.
We report the mean average precision (MAP)
and precision at top-rank 200 (precision@200), same with previous deep hashing
methods~\cite{lin2015deep,zhao2015deepsemantic,liu2016deepsupervised,liu2017deepsketchhashing}
for a fair comparison.

\subsubsection{Competitors}
We compare our deep sketch hashing model with several state-of-the-art deep hashing approaches
and for a fair comparison, we evaluate all competitors under same base network if applicable.
\noindent DLBHC~\cite{lin2015deep} replaces our dual-branch CNN-RNN network with a single-branch CNN network,
with softmax cross entropy loss used for joint feature and hashing code learning.
DSH-Supervised~\cite{liu2016deepsupervised, liuhaomiao2019ijcv} corresponds to a single-branch CNN model,
but with noticeable difference in how to model the category-level discrimination,
where pairwise contrastive loss is used based on the semantic pairing labels.
We generate online image pairs within each training batch.
DSH-Sketch~\cite{liu2017deepsketchhashing} is proposed to specifically target
on modeling the sketch-photo cross-domain relations with a semi-heterogeneous network.
To fit in our setting, we adopt the single-branch paradigm and their semantic factorization loss,
where word vector is assumed to represent the visual category. We keep other settings the same.
GreedyHash~\cite{su2018greedy} is a very recently released hashing method, using the greedy
principle to optimize discrete-constrained hashing learning by iteratively updating.

We compare with six unsupervised (Principal Component Analysis Iterative Quantization
 (PCA-ITQ)~\cite{gong2013iterative}, Locality-Sensitive Hashing (LSH)~\cite{andoni2006near},
 Spectral Hashing (SH)~\cite{weiss2009spectral}, Locality-Sensitive Hashing
 from Shift-Invariant Kernels (SKLSH)~\cite{raginsky2009locality},
 Density Sensitive Hashing (DSH)~\cite{jin2014density},
 Principal Component Analysis Hashing (PCAH)~\cite{wang2006PCAH}),
 and two supervised (Supervised Discrete Hashing~(SDH)~\cite{shen2015supervised},
 Canonical Correlation Analysis Iterative Quantization (CCA-ITQ)~\cite{gong2013iterative})
 shallow hashing methods, where deep features are fed into directly for learning.
 It's noteworthy that running each of the above eight tasks needs about $100-200$ GB memory.
 Limited by this, we train a smaller model and use $256d$ deep feature (extracted from our fusion layer) as inputs.

\subsubsection{Results and Discussions}
~\\
\noindent\textbf{Comparison against Deep Hashing Competitors}\quad
We compare our full hashing model and four state-of-the-art
deep hashing methods. Table \ref{table:compare_with_deep_baseline} shows the results for sketch hashing retrieval
under both metrics. We make the following observations: (i) Our model consistently outperforms
the state-of-the-art deep hashing methods by a significant margin,
with 6.11/8.36 and 5.50/4.79 percent improvements (MAP/Precision@200)
over the best performing competitor at 16-bit and 64-bit respectively.
(ii) The gap between our model and DLBHC suggests the benefits of combining segment-level temporal information
exhibited in a vector sketch with static pixel visual cues, the basis forming our dual-branch CNN-RNN network,
which may credit to (1) despite human tends to draw abstractly,
they do share certain category-level coherent drawing styles,
\ie, starting with a circle when sketching a sun, such that introducing additional discriminative power;
(2) CNN suffers from sparse pixel image input \cite{yu2017sketch}
but prevails at building conceptual hierarchy \cite{mahendran2016visualizing},
where RNN-based vector input brings the complements.
(iii) DSH-Supervised is unsuitable for retrieval across a large number of categories due to the incident
imbalanced input of positive and negative pairs~\cite{lin2017discriminative}.
We have also tried another very recently published pairwise similarity-preserving hashing model Deep Collaborative Discrete Hashing (DCDH)~\cite{wang2019deepsigir} as our baseline, however its performance equals to chance-performance, so that is not reported in Table~\ref{table:compare_with_deep_baseline}.
This shows the importance of metric selection under universal (hundreds of categories)
million-scale sketch hashing retrieval, where softmax cross entropy loss generally works better,
while pairwise contrastive loss hardly constrains the feature representation space and word vector can be misleading,
\ie, basketball and apple are similar in terms of shape abstraction, but pushing further away under semantic distance.
(iv) 
GreedyHash~\cite{su2018greedy} obtains unsatisfactory results mainly due to that it does not provide theoretical
clue of how the trained codes are related to data semantics~\cite{shen2019embarrassingly}.
This experimental phenomenon also evaluates the importance of semantic separability on sketch representations.

\noindent\textbf{Comparison against Shallow Hashing Competitors}\quad
In Table \ref{table:shallow_hash_method_results},
we report the performance on several shallow hashing competitors,
as a direct comparison with the deep hashing methods in Table \ref{table:compare_with_deep_baseline},
where we can observe that: (i) Shallow hashing learning generally fails to compete with joint end-to-end deep learning,
where supervised shallow methods outperform unsupervised competitors.
(ii) Under the shallow hashing learning context, deep features outperform shallow hand crafted
features by one order of magnitude.

\begin{table}[!t]
\small
\begin{center}
\caption{Retrieval time (s) per query and memory load (MB) on QuickDraw-3.8M retrieval gallery (345,000 sketches).}

\label{table:running}
\resizebox{\columnwidth}{!}{
\begin{tabular}{c || c | c | c | c}
\hline
   &16 bit &24 bit &32 bit & 64 bit   \\
\hline
\hline
Retrieval time per query (s) & 0.089 & 0.126 & 0.157 & 0.286 \\
Memory load (MB)  & 612 & 667 & 732 & 937 \\
\hline
\end{tabular}}
\end{center}


\end{table}

\begin{figure}[!t]
	\centering
	\subfigure[\scriptsize{Precision-Recall curves @$16$ bits}]{
		\label{fig:ps_16bits}
		\includegraphics[width=0.23\textwidth]{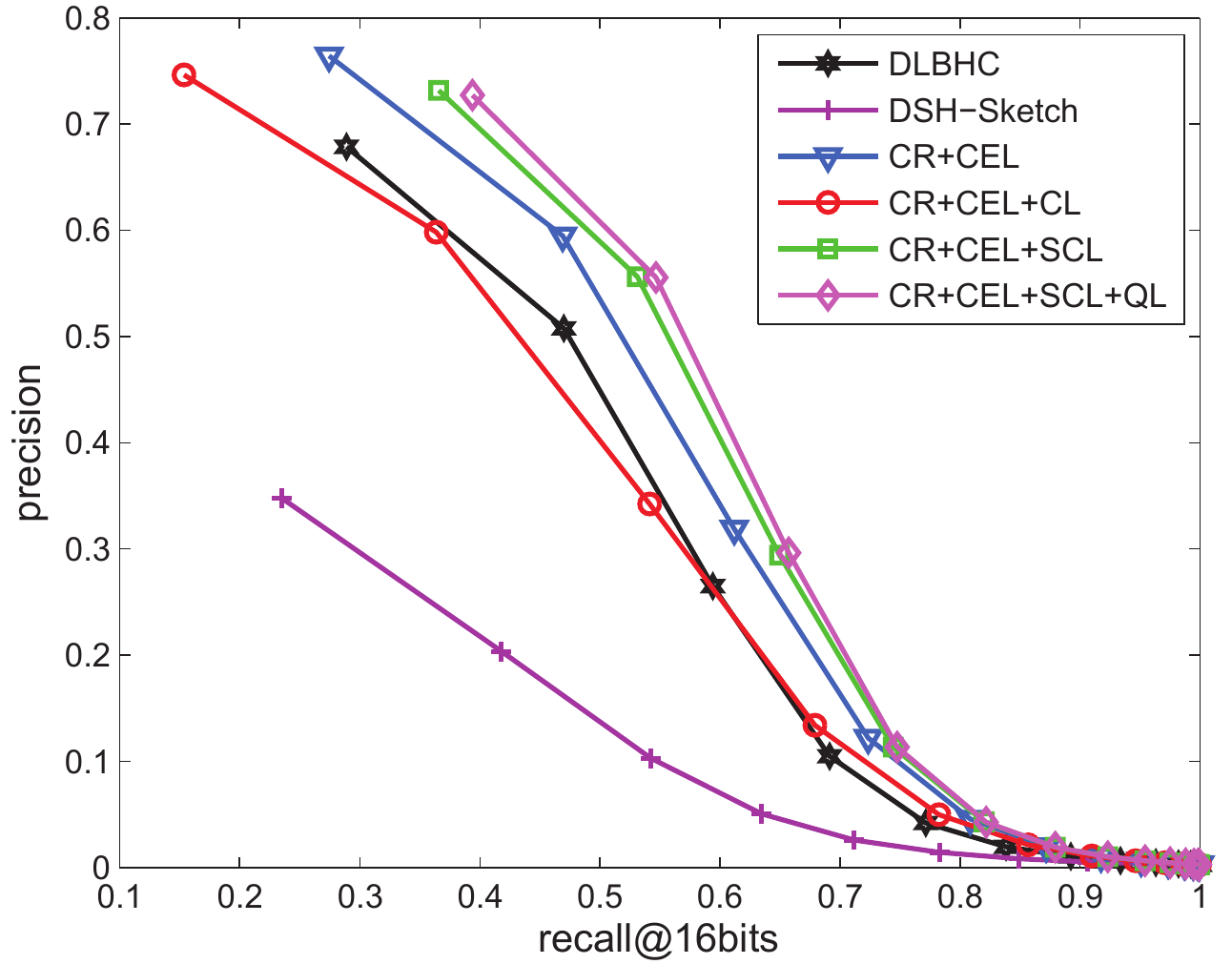}}
	\subfigure[\scriptsize{Precision-Recall curves @$24$ bits}]{
		\label{fig:ps_24bits}
		\includegraphics[width=0.23\textwidth]{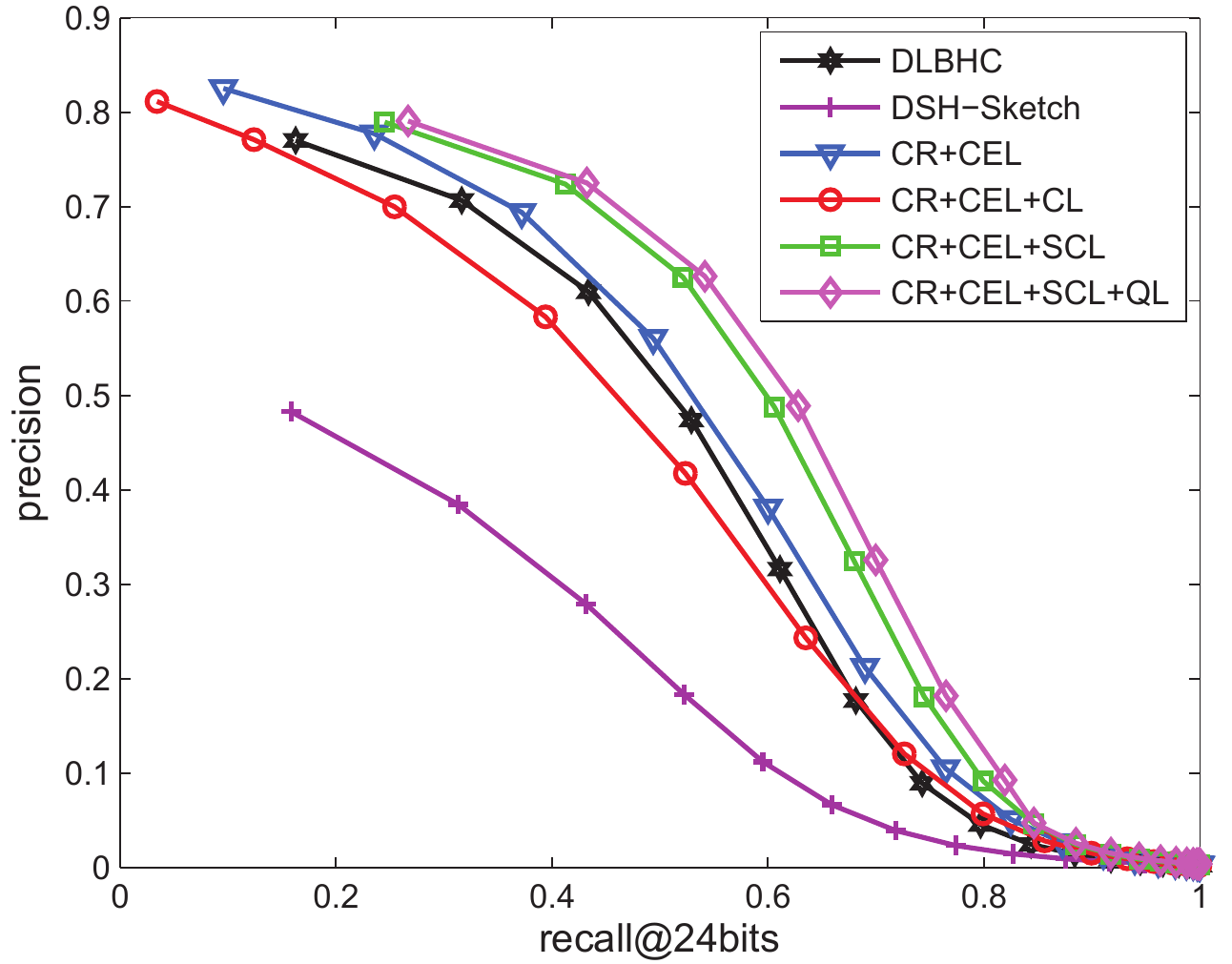}}
    \subfigure[\scriptsize{Precision-Recall curves @$32$ bits}]{
		\label{fig:ps_32bits}
		\includegraphics[width=0.23\textwidth]{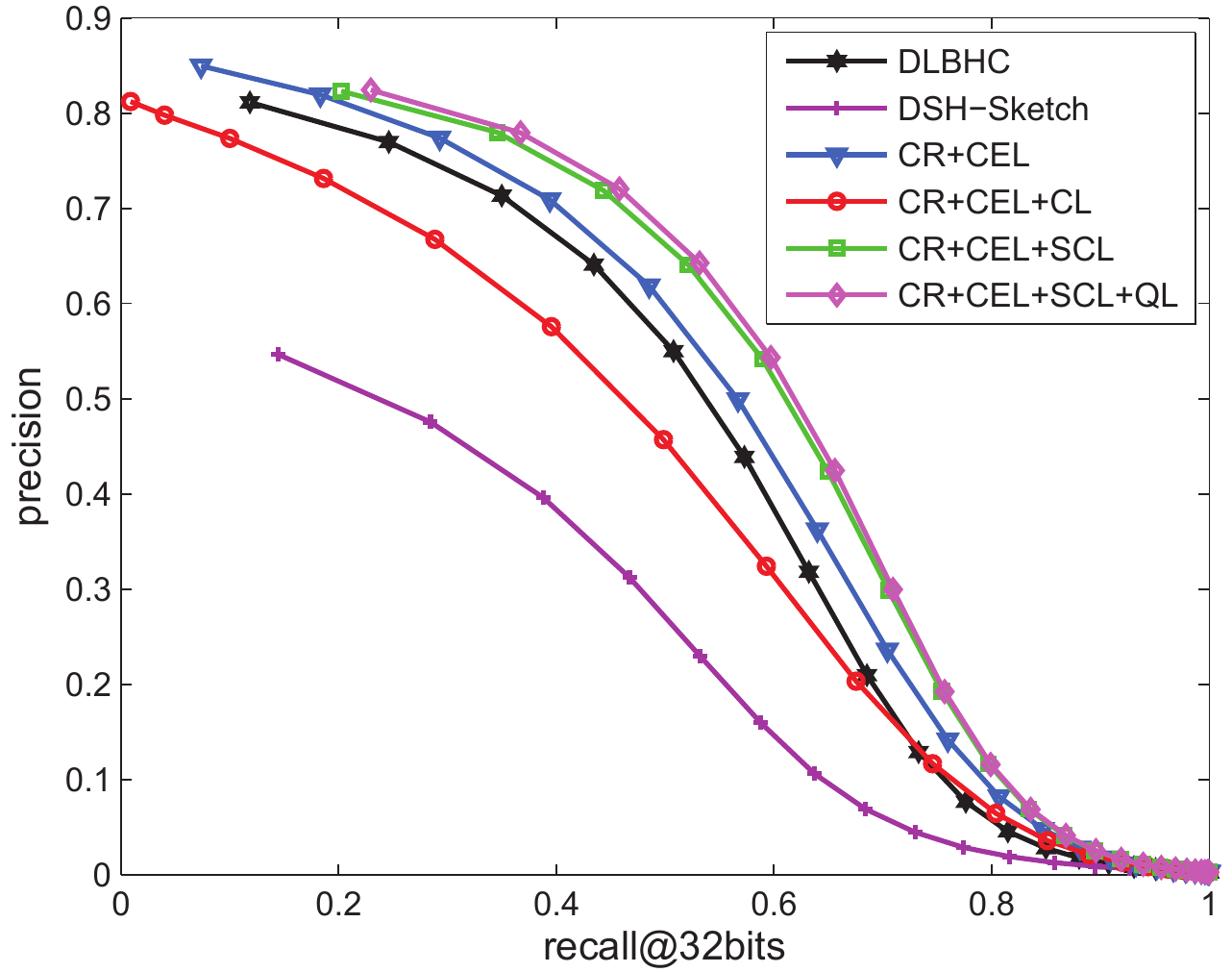}}
    \subfigure[\scriptsize{Precision-Recall curves @$64$ bits}]{
		\label{fig:ps_64bits}
		\includegraphics[width=0.23\textwidth]{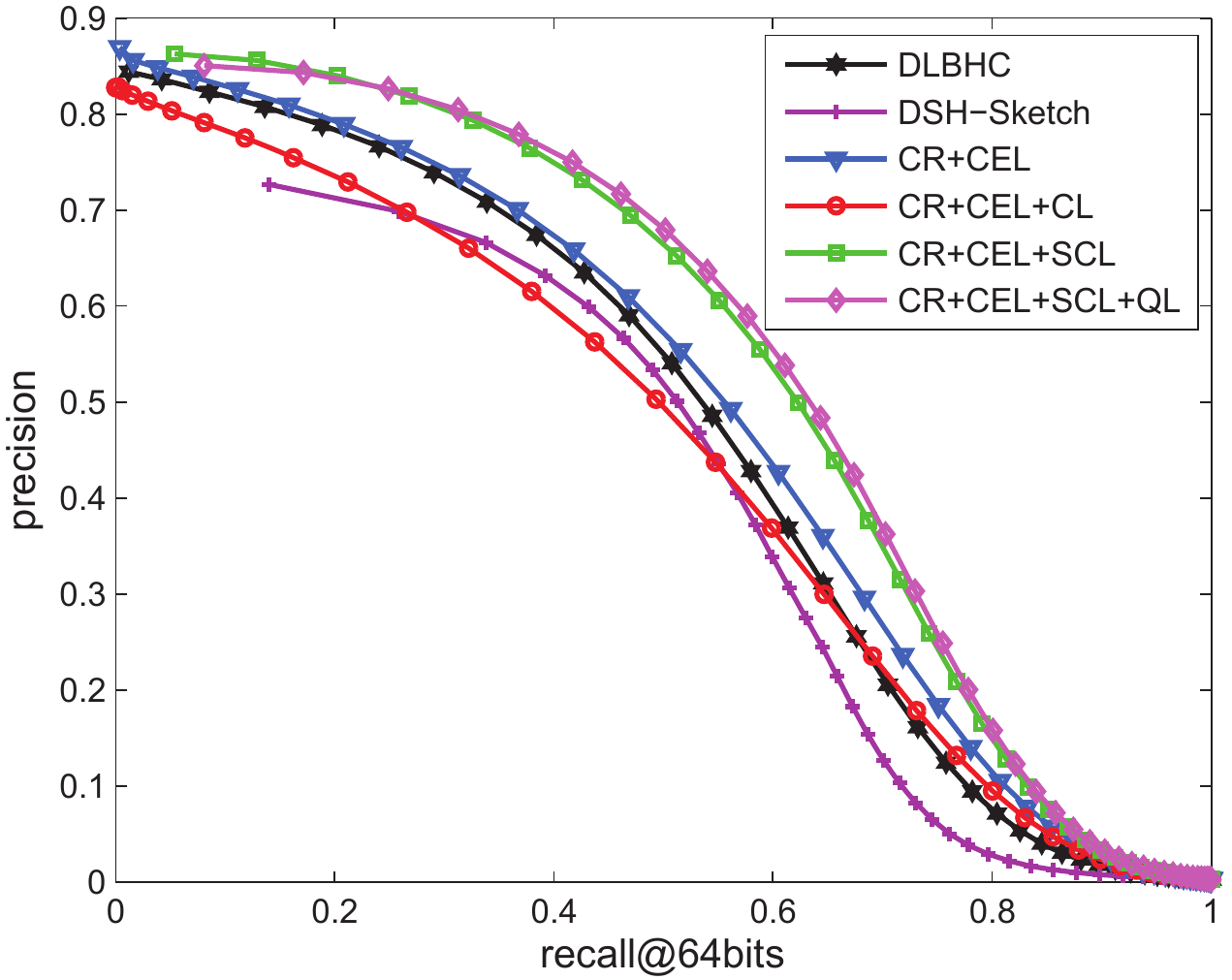}}
	\caption{Precision recall curves on QuickDraw-3.8M retrieval gallery. Best viewed in color.}
	\label{fig:precision_recall_curves}
\end{figure}

\noindent\textbf{Component Analysis}\quad
We have also evaluated the effectiveness of different components of our model
in Table~\ref{table:compare_with_deep_baseline}.
Specifically, we combine our CR network training with different loss combinations,
including softmax cross entropy loss (CR+CEL), softmax cross entropy plus common center loss (CR+CEL+CL),
softmax cross entropy plus sketch center loss (CR+CEL+SCL), softmax cross entropy plus sketch center loss plus
quantization loss (CR+CEL+SCL+QL), which arrives our full hashing model.
We find that with cross entropy loss alone
under our dual-branch CNN-RNN network suffices to outperform best competitor, where by adding sketch center loss
and quantization loss further boost the performance. It's noteworthy that adding common center loss harms
the performance quite significantly, validating our sketch-specific center loss design.
In Figure \ref{fig:precision_recall_curves}, we plot the precision-recall curves for all
above-mentioned methods under $16$, $24$, $32$ and $64$ bit hashing codes respectively,
which further matched our hypothesis.

%

\begin{table}[!t]
\footnotesize
\begin{center}
\caption{Comparison with the state-of-the-art methods and our hashing model variants on sketch recognition task on QuickDraw-3.8M retrieval gallery.}
\label{table:classification_accuracies}
\resizebox{0.4\textwidth}{!}{
\begin{tabular}{l || c}
\hline
Model    & Classification Accuracy \\
\hline
\hline
Sketch-a-Net~\cite{yu2017sketch} & 0.6871 \\
ResNet-50 ~\cite{he2016deep} &  0.7864  \\
CR + CEL &          0.7949                  \\
CR + CEL + SCL  &     \textbf{0.8051}         \\
\hline
\end{tabular}
}
\end{center}

\end{table}

\begin{table}[!t]
\begin{center}
\caption{Statistic analysis for distances in the feature space of QuickDraw-3.8M under our hashing model variants.
$d_1$ and $d_2$ denote intra-class distance and inter-class distance, respectively.
``CR'' denotes our sketch-specific dual-branch CNN-RNN network.}

 \label{table:distance_statistic_analysis}
\resizebox{\columnwidth}{!}{
\begin{tabular}{c|c||c|c|c|c}
\hline
\multicolumn{2}{c||}{~}  & CR+CEL & CR+CEL+CL & CR+CEL+SCL & \tabincell{c}{CR+CEL+SCL+QL \\ (Our Full Hashing Model)} \\
\hline \hline
\multirow{4}{*}{16 bits} & $d_1$             & 0.7501 & 0.5297 & 0.5078 & 0.5800 \\
                       & $d_2$             & 4.9764 & 3.2841 & 4.2581 & 4.8537 \\
                       & $d_1 / d_2$       & 0.1665 & 0.1721 & \textbf{0.1257} & 0.1290 \\
                       & MAP               & 0.5969 & 0.5567 & 0.6016 & \textbf{0.6064}\\
\hline\hline
\multirow{4}{*}{24 bits} & $d_1$             & 1.2360 & 0.8285 & 0.6801 & 0.8568 \\
                       & $d_2$             & 6.1266 & 4.0388 & 5.0221 & 6.2243\\
                       & $d_1 / d_2$       & 0.2017 & 0.2051 & \textbf{0.1354} & 0.1377\\
                       & MAP               & 0.6196 & 0.5856 & 0.6374 & \textbf{0.6388}\\
\hline\hline
\multirow{4}{*}{32 bits} & $d_1$             & 2.0066 & 1.5124 & 1.0792 & 1.2468 \\
                       & $d_2$             & 8.9190 & 7.3120 & 7.5340 & 8.6675 \\
                       & $d_1 / d_2$       & 0.2250 & 0.2068 & \textbf{0.1432} & 0.1439 \\
                       & MAP               & 0.6412 & 0.5911 & 0.6473 & \textbf{0.6521} \\
\hline\hline
\multirow{4}{*}{64 bits} & $d_1$             & 4.7040 & 3.5828 & 1.6109 & 2.5231\\
                       & $d_2$             & 15.4719 & 14.1112 & 11.6815 & 17.6179\\
                       & $d_1 / d_2$       & 0.3040 & 0.2539 & \textbf{0.1379} & 0.1432\\
                       & MAP               & 0.6525 & 0.6136 & 0.6767 & \textbf{0.6791} \\
\hline
\end{tabular}}

\end{center}

\end{table}

\begin{figure*}[!t]
\begin{center}
\includegraphics[width=\textwidth]{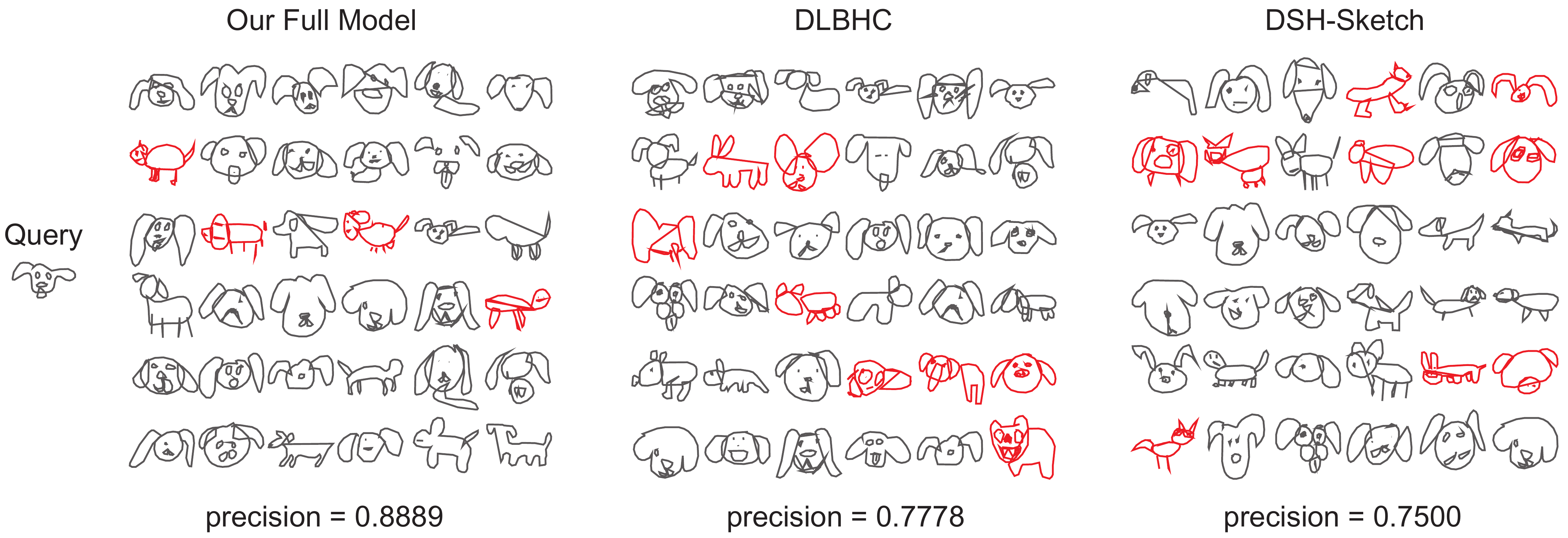}
\end{center}
   \caption{Qualitative comparison of top $36$ retrieval results of our model and the state-of-the-art deep hashing methods for query (dog) at $64$ bits on QuickDraw-3.8M retrieval gallery.
   Red sketches indicate false positive sketches. The retrieval precision is obtained
   by computing the proportion of true positive sketches. Best viewed in color. }
\label{fig:qualitative}
\end{figure*}

\noindent\textbf{Further Analysis on Sketch Center Loss}\quad
To statistically illustrate the effectiveness of
our sketch center loss, we calculate the average ratio of the intra-class distance $d_1$ and inter-class distance $d_2$,
termed as $d_1/d_2$, among our $345$ training categories.
A lower value of such score indicates a better feature representation learning,
since the objects within the same category tend to cluster tighter and push further away with those of different
semantic labels, as forming a more discriminative feature space. In Table \ref{table:distance_statistic_analysis},
we witness significant improvement on the category structures brought by the sketch
center loss across all hashing length setting (CR+CEL vs. CR+CEL+SCL), where on contrary,
common center even undermines the performance (CR+CEL vs. CR+CEL+CL), which in accordance
with what we've observed in Table \ref{table:compare_with_deep_baseline}.

\noindent\textbf{Qualitative Evaluation}\quad
In Figure \ref{fig:qualitative}, we qualitatively compare our full hashing model
with DLBHC \cite{lin2015deep} and DSH-Sketch \cite{liu2017deepsketchhashing} on the dog category.
It's interesting to observe (i) how our model makes less semantic mistakes;
(ii) how our mistake is more reasonably understandable, \ie,
sketch is confusing in itself in most of our falsely-retrieved sketches,
while in other methods some clear semantic errors take place (\eg, pigs and rabbits).

\noindent\textbf{Running Cost}\quad
We report the running cost as retrieval time (s)
per query and memory load (MB) on QuickDraw-3.8M retrieval gallery (345,000 sketches)
in Table \ref{table:running}, which even on million-scale can still achieve real-time retrieval performance.

\noindent\textbf{Generalization to Sketch Recognition}\quad
To validate the generality of our sketch-specific design,
we apply our dual-branch CNN-RNN network to sketch recognition task,
by directly use a $2048D$ fully connected layer as fusion
layer
before the $345$-way classification layer.
We compare with two state-of-the-art classification networks
-- Sketch-a-net \cite{yu2017sketch} and ResNet-50 \cite{he2016deep},
where all these sketch recognition experiments are evaluated on the QuickDraw-3.8M retrieval gallery set.
We demonstrate the results in Table \ref{table:classification_accuracies},
where following conclusions can be drawn: (i) Exploiting the sketching temporal orders is important,
and by combining the traditional static pixel representation, more discriminative power
is obtained ($79.49\%$ vs. $68.71\%$). (ii) Sketch center loss generalizes to sketch recognition task,
bringing additional benefits.

\noindent\textbf{Generalization to Zero-Shot Sketch Hashing}\quad
We randomly pick $20$ categories from QuickDraw-3.8M and exclude them from training. 
We follow the same experimental procedures on 32bit hash codes and report the MAP performance on the unseen categories. 
As reported in Table~\ref{table:zeroshot_sketch_hashing}, under such challenging seen-unseen split, our method's MAP of $0.7547$ outperforms that of DLBHC ($0.7094$) and DSH-Sketch ($0.5334$), by a clear margin.

\begin{table}[!t]
\footnotesize
\begin{center}
\caption{Comparison on zero-shot sketch hashing on QuickDraw-3.8M.}
\label{table:zeroshot_sketch_hashing}
\resizebox{\columnwidth}{!}{
\begin{tabular}{l || c}
\hline
Model    &  Mean Average Precision \\
\hline
\hline
DLBHC~\cite{lin2015deep} &  0.7094 \\
DSH-Sketch~\cite{liu2017deepsketchhashing} &  0.5334  \\
 CR+CEL+SCL+QL (Our Full Hashing Model)  &     \textbf{0.7547}         \\
\hline
\end{tabular}
}
\end{center}

\end{table}

\begin{table}[!t]
\small
\begin{center}
\caption{Benchmark of the state-of-the-art CNNs on our large-scale edge-map dataset.}
\label{table:edgemap_benchmark}
\resizebox{0.9\columnwidth}{!}{
\begin{tabular}{l || c}
\hline
 Model & Validation Accuracy  \\
\hline
\hline
MobileNet V1~\cite{howard2017mobilenets} & 0.6108  \\
MobileNet V2~\cite{sandler2018mobilenetv2} & 0.5971   \\
ResNet-18~\cite{he2016deep} & 0.5943 \\
ResNet-34~\cite{he2016deep} &  0.5931 \\
ResNet-50~\cite{he2016deep} &   0.5896 \\
ResNet-152~\cite{he2016deep} &  0.5829 \\
DenseNet-121~\cite{huang2017densely} & 0.6153 \\

DenseNet-161~\cite{huang2017densely} & 0.5518 \\
DenseNet-169~\cite{huang2017densely} & 0.5177 \\
DenseNet-201~\cite{huang2017densely} &  0.5514 \\

GoogLeNet Inception V3~\cite{szegedy2016rethinking}   &         0.6237        \\

VGG-11~\cite{simonyan2014very}        &         0.6132        \\
\hline
\end{tabular}
}
\end{center}

\end{table}

\begin{table*}[!t]
\caption{Comparison with the state-of-the-art photo zero-shot learning approaches on 145 classes of QuickDraw-3.8M validation set.
}

\label{table:zero-shot-acc}
\tiny
\begin{center}

\resizebox{\textwidth}{!}{
\begin{tabular}{l || c| c| c| c || c| c |c |c}
\hline
  \multirow{2}{*}{Model} & \multicolumn{4}{c||}{ZSL Accuracy} &  \multicolumn{4}{c}{GZSL Accuracy} \\
  \cline{2-9}
 & hit@1 & hit@5 & hit@10 & hit@20 &  hit@1 & hit@5 & hit@10 & hit@20 \\
\hline
\hline
 SAE (sketch$\rightarrow$word vector)~\cite{kodirov2017semantic}          & 0.0817 & 0.2278& 0.3413& 0.5013 &   0.0011 & 0.0470 & 0.1105 & 0.2314  \\
 SAE (sketch$\leftarrow$word vector)~\cite{kodirov2017semantic}          & 0.1056 & 0.2751& 0.3940&  0.5471  &  0.0085 & 0.1387 & 0.2555 &  0.3986 \\
 DEM~\cite{lizhang2017zero} &  0.1224 & 0.2818 & 0.3951&  0.5347    &   0.0312 & 0.2150 & 0.3398 & 0.4953  \\
 DEM~(sketch $\rightarrow$ word vector)~\cite{lizhang2017zero} & 0.0198 & 0.0601& 0.1233 & 0.2314 &   0.0058 & 0.0292 & 0.0548 &  0.1069  \\
 RELATION NET~\cite{sung2018learning} & 0.0968 & 0.2828&0.4076 &  0.5654 &   0.0070 & 0.1185 & 0.2184 &  0.3562 \\
 AREN~\cite{xie2019attentive} & 0.0156 & 0.0781 & 0.1328 &  0.2188 &   0.0025 & 0.0175 & 0.0199 &  0.0430 \\
 \hline
 \hline
 Our Full SZSR Model & \textbf{0.2148} & \textbf{0.5031} & \textbf{0.6363} & \textbf{0.7589}  &     \textbf{0.0756}& \textbf{0.3360}& \textbf{0.4831} & \textbf{0.6416}\\
\hline
\end{tabular}
}
\end{center}

\end{table*}

\begin{table*}[!t]
\normalsize
\begin{center}
\caption{Ablation study for our SZSR model variants on 145 classes of QuickDraw-3.8M validation set. ``CR'' denotes the sketch-specific dual-branch CNN-RNN network.}
\label{tab:zero-shot-ablation}
\resizebox{\textwidth}{!}{
\begin{tabular}{c|c|c || c | c | c | c || c | c | c | c}
\hline
 \multirow{2}{*}{Model} & \multirow{2}{*}{SE Subnet}& \multirow{2}{*}{Embedding Direction} & \multicolumn{4}{c||}{ZSL Accuracy} & \multicolumn{4}{c}{GZSL Accuracy}\\
 \cline{4-11}
& & & hit@1  & hit@5 &  hit@10& hit@20&       hit@1 & hit@5 & hit@10 & hit@20\\
\hline
\hline
 \multirow{5}{*}{Our Ablation Models}    & CNN & sketch $\rightarrow$ edge-map  & 0.0405  & 0.0845 & 0.1312& 0.2215   &  0.0049&  0.0477& 0.0898 & 0.1281\\
                                     & RNN & sketch $\rightarrow$ edge-map  & 0.0384  & 0.1195 & 0.1903 & 0.3003 &     0.0087& 0.0637& 0.1043 &0.1691\\
                                     & CR  & sketch $\rightarrow$ edge-map  & 0.0435  & 0.1361 &  0.2198 &0.3408 &     0.0097& 0.0668& 0.1169 & 0.1932\\
                                     & CNN & sketch $\leftarrow$ edge-map  & 0.1919   & 0.4732 & 0.6062 & 0.7395&     0.0638& 0.2947 & 0.4441 & 0.5936\\
                                     & RNN & sketch $\leftarrow$ edge-map  & 0.1815  & 0.4516 & 0.5969& 0.7483&     0.0554& 0.2632& 0.4040 & 0.5664 \\
 \hline
 \hline
 Our Full SZSR Model                      & CR  & sketch $\leftarrow$ edge-map   & \textbf{0.2148}  & \textbf{0.5031} &\textbf{0.6363} &\textbf{0.7589} &     \textbf{0.0756}& \textbf{0.3360}& \textbf{0.4831} & \textbf{0.6416}\\
\hline
\end{tabular}
}
\end{center}

\end{table*}

\subsection{Zero-Shot Recognition}

\subsubsection{Datasets and Settings}
We randomly select $200$ classes from Google QuickDraw dataset as our \textit{seen} classes,
using the remaining $145$ classes as our \textit{unseen} classes.
In training, we use the selected $200$ classes of 
 QuickDraw-3.8M
training set,
\ie, totally $1, 800, 000$ sketches ($200 \times 9000$)
.
In testing, our zero-shot accuracy is calculated on the
selected $145$ \textit{unseen} classes of QuickDraw-3.8M validation set,
\ie, totally $145, 000$ sketches ($145 \times 1000$) for zero-shot testing.

\subsubsection{Implementation Details}
Our sketch encoding (SE) subnet is implemented
by the sketch-specific dual-branch CNN-RNN network
,
and all the detailed configurations are the same with the description in above section.
\textcolor{black}{We empirically use $4096D$ fully connected layer with Rectified
Linear Unit (ReLU) activation as the fusion layer for our sketch zero-shot recognition task.}

Our edge-map vector encoding (EVE) subnet is implemented by
two fully-connected layers 
($1280D \rightarrow 2048D$, $2048D \rightarrow 4096D$) using ReLU as activation.

We randomly split our edge-map dataset into $263,655$
and $26,626$ edge-maps for training and validation, respectively.
Moreover we benchmark state-of-the-art CNN networks on our edge-map dataset
, and results are reported in Table~\ref{table:edgemap_benchmark} (all the CNNs are trained from scratch).
In our following SZSR experiments,
we choose MobileNet V$2$ trained on our edge-map training set as our feature extractor for faster calculation,
and extract features for our edge-map training set.
We calculate a mean feature vector~($1280D$) for each edge-map class,
thus finally we can obtain a $345 \times 1280D$ edge-map vector set.

All our zero-shot recognition experiments are implemented in PyTorch\footnote{\url{https://pytorch.org/}}, and run on a single TITAN X GPU.
We use RMSprop optimizer for all our training stages.
We report the zero-shot learning (ZSL) accuracy and general zero-shot learning (GZSL) accuracy, same
with previous zero-shot learning works~\cite{kodirov2017semantic,xian2018zero} 
for
a fair comparison.


\subsubsection{Competitors}
To our knowledge, no existing methods can be compared directly.
Therefore, we have to compare with the state-of-the-art photo zero-shot learning methods, mainly including
semantic autoencoder~(SAE)~\cite{kodirov2017semantic},
deep embedding model~(DEM)~\cite{lizhang2017zero},
RELATION NET~\cite{sung2018learning}, Attentive Region Embedding Network for zero-shot learning (AREN)~\cite{xie2019attentive}.
These competitors have performed
well using word vector or attribute vector as semantic knowledge
,
thus we evaluate them based on word vector,
and replace their visual domain with sketch, keeping the remaining settings the same for a fair comparison.
Moreover, in order to demonstrate the different effects of different embedding spaces,
we run SAE and DEM in both of ``visual to semantic'' and ``semantic to visual'' modes.

\subsubsection{Results and Discussions}
~\\
\noindent\textbf{Comparison with Photo ZSL Models}\quad
As illustrated in Table~\ref{table:zero-shot-acc}, our proposed model achieves $0.2148$ ZSL accuracy~(hit@1),
whilst the highest baseline performance is $0.1224$~(hit@1).
This obvious gap illustrates the advantage of our sketch-specific design combination,
\ie, (i) using dual-branch CNN-RNN network to perform sketch feature representation,
(ii) using edge-map vectors as semantic knowledge to conduct domain alignment.
For GZSL accuracy, our proposed model also outperforms all the baselines by a large margin.

In Table~\ref{table:zero-shot-acc}, each baseline obtains different 
ZSL
accuracies based on different embedding directions ($0.0817$ vs. $0.1056$, $0.1224$ vs. $0.0198$).
This demonstrates that choosing a reasonable embedding direction is important for zero-shot embedding. For GZSL setting, we can observe similar phenomenon in Table~\ref{table:zero-shot-acc}.

\noindent\textbf{Ablation Study}\quad
We conduct the ablation study for our proposed embedding model by combining different embedding directions
with different components of 
the sketch-specific
CNN-RNN network.
In Table~\ref{tab:zero-shot-ablation}, we observe that, under ZSL setting:
(i) All the models \textcolor{black}{using sketch feature domain as embedding space} outperform their corresponding models
using edge-map 
vector domain as embedding space
by a large margin~($0.1919$ vs. $0.0405$, $0.1815$ vs. $0.0384$, $0.2148$ vs. $0.0435$). 
In another word, our deep embedding model is always sensitive to the embedding direction, based on different SE subnets~(\ie, CNN, RNN, CNN-RNN). 
This phenomenon shares accordance with the hubness issue illustrated in~\cite{lizhang2017zero}.
(ii)
For both embedding directions, the sketch-specific dual-branch CNN-RNN network outperforms its single branches.
(iii)
Compared with all the baselines reported in Table~\ref{table:zero-shot-acc},
our CNN SE subnet based and RNN SE subnet based ablation models nearly obtain their double classification accuracies,
when we embed edge-map prototypes into sketch feature domain.
In Table~\ref{tab:zero-shot-ablation}, we observe similar results under GZSL setting.

Based on these observations, we can draw several conclusions:
(i) Sketches are different from photos. In zero-shot recognition scenario,
sketch can use semantic knowledge extracted from edge-map domain to conduct domain alignment,
which has not been exploited in zero-shot learning field.
For SZSR, semantic knowledge extracted from edge-map domain is more reasonable than word vector based semantic knowledge.
(ii) 
For sketch zero-shot recognition embedding model, choosing a reasonable embedding space is important.
(iii) 
For sketch zero-shot recognition, the sketch-specific dual-branch CNN-RNN network
also provides better feature representation than both of single CNN and single RNN networks.

\section{Conclusion}
\label{sec:conclusion}
In this paper, we aim to study learning semantic representations for million-scale free-hand sketches to explore the unique traits of large-scale sketches that were otherwise under-studied in the prior art.
{By grasping the intrinsic traits of free-hand sketches, we propose a dual-branch CNN-RNN architecture to learn high-level semantic representations for free-hand sketches, where utilize CNN to extract abstract visual
concepts and RNN to model human sketching temporal orders, respectively.
This dual-branch architecture can provide discriminative feature representations for various sketch-oriented tasks.}
Based on this architecture, we further explore learning the sketch-oriented semantic representations in two challenging yet practical settings, \ie, hashing retrieval and zero-shot recognition on million-scale sketches.
Specifically, we use our dual-branch architecture as a universal representation framework to design two sketch-specific deep models.
For sketch hashing retrieval,
we propose a novel hashing loss that accommodates the abstract nature of sketches.
Our hashing model outperforms the state-of-the-art shallow and deep alternatives,
and yields superior generalization performance when re-purposed for sketch recognition.
For sketch zero-shot recognition,
we propose to use edge-map based semantic vectors as semantic knowledge for domain alignment, and 
we collect a large-scale edge-map dataset that covers all the $345$ sketch classes of QuickDraw dataset to obtain high-quality edge-maps.
By using our proposed edge-map vector and CNN-RNN architecture, 
we design a deep embedding model for large-scale sketch zero-shot
recognition, which outperforms the state-of-the-art zero-shot learning models.

{\small
\bibliographystyle{IEEEtran}
\bibliography{egbib}
}

\end{document}